# Figurative Archive: an open dataset and web-based application for the study of metaphor

Maddalena Bressler*, Veronica Mangiaterra*, Paolo Canal, Federico Frau, Fabrizio Luciani, Biagio Scalingi, Chiara Barattieri di San Pietro, Chiara Battaglini, Chiara Pompei, Fortunata Romeo, Luca Bischetti¶, & Valentina Bambini¶

Laboratory of Neurolinguistics and Experimental Pragmatics (NEPLab), University School for Advanced Studies IUSS, Pavia, Italy

* Contributed equally as co-first authors
¶ Contributed equally as co-last authors

**Corresponding author**: Valentina Bambini, Laboratory of Neurolinguistics and Experimental Pragmatics (NEPLab), University School for Advanced Studies IUSS, Palazzo del Broletto
Piazza della Vittoria 15, 27100, Pavia, Italy. Tel: +39 0382 375811, fax: +39 0382 375899
E-mail: valentina.bambini@iusspavia.it

**Abstract**
Research on metaphor has steadily increased over the last decades, as this phenomenon opens a window into a range of linguistic and cognitive processes. At the same time, the demand for rigorously constructed and extensively normed experimental materials increased as well. Here, we present the *Figurative Archive*, an open database of 996 metaphors in Italian enriched with rating and corpus-based measures (from familiarity to semantic distance and preferred interpretations), derived by collecting stimuli used across 11 studies. It includes both everyday and literary metaphors, varying in structure and semantic domains, and is validated based on correlations between familiarity and other measures. The *Archive* has several aspects of novelty: it is increased in size compared to previous resources; it offers a measure of metaphor inclusiveness, to comply with recommendations for non-discriminatory language use; it is displayed in a web-based interface, with features for a customized consultation. We provide guidelines for using the *Archive* to source materials for studies investigating metaphor processing and relationships between metaphor features in humans and computational models.

**Keywords**
Figurative language; Metaphor; Pragmatics; Experimental Pragmatics; Neuropragmatics; Psycholinguistics

## Background

Typically defined as a language use where one thing is described in terms of something else that is conceptually very different (as in the case of *This archive is a gem*), metaphor is a phenomenon that straddles the border between rhetoric, philosophy, linguistics, and psychology[1]. In the last few decades, metaphor research has expanded well beyond classic literary studies, entering the fields of psycholinguistics, neurolinguistics, and cognitive neuroscience more broadly[2–4]. Bibliometric studies[5–7] indicate a stable upward trend in metaphor research over time, with a marked rise in the early 2010s due to the introduction of experimental methods[8,9]. One of the reasons behind such a growing interest is that metaphor offers a window into different cognitive processes. It is used, for instance, to investigate inferential mechanisms within the field of Experimental Pragmatics and neuropragmatics[10–13], to explore embodied and simulation processes within the field of Cognitive Linguistics and Grounded Cognition[14], to test abstraction in neurotypical as well as clinical samples[1,15,16], to study high-level language acquisition in L1 and L2[17,18], up to aesthetic appreciation in neurocognitive poetics[19,20].

One finding that emerges clearly from the literature above is that each metaphor is a multifaceted object, with many attributes affecting its processing[21]. These encompass metaphor familiarity, which might reduce processing efforts[22–24] and the degree of sensorimotor reenactment[14,25], concreteness of metaphorical expressions, with different patterns of acquisition and decay in the lifespan for more concrete vs. more abstract metaphors[26–28], aptness[29], which eases comprehension and favors the categorization processes[30], as well as a number of word-level semantic features[31,32]. Such evidence has stimulated a large debate over the distinctiveness of the metaphor features[33] and, in general, has elucidated that metaphors elicit different behavioral and brain response patterns depending on their specific linguistic characteristics[34].

Given the scenario above, experimental research on metaphor requires a great deal of attention when constructing and selecting the testing material. In most cases, each study includes a specific phase devoted to crafting the metaphors and collecting *de novo* rating measures from participants. This, however, is not only time-consuming but also hampers reproducibility. In an attempt to overcome these limitations, a number of datasets enriched with human ratings have been published in recent decades, especially for English. Starting from the pioneering work of Katz et al. (1988)[35], which comprises 260 nonliterary (i.e., of use in everyday life and ordinary language) and 204 literary metaphors with 10 dimensions, a portion of which (n = 50) was recently renormed by Campbell & Raney (2016)[36], other datasets include those of Cardillo et al. (2010)[37] and Cardillo et al. (2017)[38], respectively with 280 and 120 metaphors and 10 measures, Roncero & de Almeida (2015)[39], with 84 metaphors and seven measures, and Thibodeau et al. (2018)[33], with 36 metaphors rated for five dimensions. Sparse and lower-scale efforts to create datasets in other languages were conducted, for instance, for German[40,41], Italian[42], Dutch[43], and Serbian[44], as well as in other language families such as Chinese[45,46], also in a cross-language perspective[47,48].

The *Figurative Archive* presented here follows in the trail of providing an open dataset of Italian metaphors with ratings for future research. Capitalizing on more than 10 years of psycholinguistic and neurolinguistic investigation on metaphor processing conducted by our research group[24,27,49], we gathered metaphors and corresponding rating and corpus-based measures from 11 individual studies, some published (seven) and some currently unpublished (four), standardized and validated them via extensive correlational analysis. The *Figurative Archive* currently includes two modules. The 464 items of the *Everyday Metaphors* module are intended to offer a resource for investigating metaphors that occur in ordinary language in different forms. The available measures, which span from familiarity (available for almost 100% of the corpus) to preferred interpretations (available for 27% of the corpus) to body relatedness (available for 14% of the corpus), show a substantial degree of variation, allowing for investigating specific features of metaphorical language. Moreover, the whole *Everyday Metaphors* module was complemented with a *de novo* collected dimension that has never been explored before, namely inclusiveness. In doing metaphor research over more than a decade timeframe, we have experienced a change in speakers' sensitivity to metaphors' discriminatory value, with participants starting, in debriefing sessions, to report low acceptance of certain metaphors, especially those referring to bodily attributes. Such a change matches the current attention at the societal level for inclusive language[50]. This aspect, however, has never been empirically measured in metaphor research. Hence, we developed an *ad hoc* questionnaire and used its outcome to complement each item with an indicator of possible discriminatory interpretations. The *Literary Metaphors* module is intended to offer a dataset of 532 original metaphors extracted from Italian literary texts, centered around classical domains such as emotions (e.g., love), natural elements (e.g., fire), physical locations (e.g., river) and events (e.g., storm). The values available for the literary metaphors (mostly corpus-based) are sufficiently distributed to make the dataset useful for exploring the role of creativity and poetic aspects.

Moreover, we organized all data both in a Zenodo repository and in an online searchable platform developed for easy navigation and customized search. The web interface was designed to offer easy and flexible consultation at different levels. In addition to displaying the 996 items and their characteristics, it allows users to constrain the search by selecting specific metaphorical terms (such as topics or vehicles) or ranges of values for the different properties. Within each module, the interface also provides two interactive tabs for the evaluation of the distribution of values and associations between measures across the dataset. Further detailed information is described in the Data Records and Usage Notes sections.

The *Figurative Archive* can promote metaphor research in several ways. As a first, most obvious advantage, it offers a set of readily usable and extensively described metaphors, mostly paired with literal counterparts, thereby reducing the time required for experiment implementation. In this light, the variety of types included in the dataset makes the *Figurative Archive* a valuable resource for research on different aspects of metaphor. Second, the *Archive* encourages reproducible research in metaphor studies, both when addressing the neurocognitive effects investigated in the original studies from which the metaphors were extracted and when serving as a shared source of material for future studies. Third, the plethora of attributes included in the *Archive* allows for systematic and large-scale investigations into the properties of metaphor, their relationships, and their impact on processing, which is still a matter of lively debate[28,33,51,52]. Fourth, it may promote the systematic testing of figurative language abilities of Large Language Models (LLMs)[53–58]. For instance, the *Archive* might serve as a base to construct benchmarks for Italian, aligning with the rising need for resources in languages other than English[59]. Also, the *Archive* may help mitigate the pitfalls of over-reliance on English[60] in metaphor research. Granted that the *Archive* contains Italian metaphors and that metaphors cannot easily be mapped from one language to another[21], some metaphorical expressions show a considerable degree of stability across languages[61]. In this vein, the *Archive* provides the English translations of the Italian metaphorical expressions, to ensure accessibility to non-Italian-speaking readers, and enables users to search for metaphors associated with specific topics or vehicles, which may be more easily translatable into English or other languages. Our plan for the future is to continue expanding the data collection by contributing new datasets ourselves and by encouraging colleagues worldwide to develop parallel or joint initiatives, to unravel the interplay between biological and cultural roots behind metaphors.

## Methods

### *Everyday Metaphors*

The *Everyday Metaphors* module of the *Figurative Archive* comprises 464 unique metaphorical expressions in Italian (405, 87.28%, paired with a literal counterpart) pooled from nine studies conducted by members of the NEPLab (https://www.neplab.it/). A unique alphanumeric ID was assigned to each metaphorical expression based on the chronological order of the original studies. The dataset features various types of metaphorical expressions, including nominal predicative metaphors (e.g., *That lawyer is a shark*), nominal metaphors in word pairs (e.g., *lake* - *crystal*), and predicate metaphors (e.g., *Luigi pushes forward through life's problems*), with indication – for each expression – of the topic (i.e., the subject of the metaphor, e.g., *lawyer* in the first example above) and the vehicle (i.e., the term used to convey the metaphorical meaning, e.g., *shark* in the first example above). A literal English translation is given for each metaphorical item, for accessibility purposes. For some of the predicate metaphors in the study by Frau et al. (2025)[62], a more idiomatic translation is provided as well, to aid the understanding of the meaning of the expression. Each metaphorical item is accompanied by a set of relevant measures, either obtained through rating tasks (familiarity, meaningfulness, difficulty, physicality, mentality, aptness, body relatedness, imageability, metaphoricity, cloze probability, entropy, number of interpretations, and strength of interpretation) or corpus-based (semantic distance between topics and vehicles, length, frequency, and concreteness of both topics and vehicles, the latter henceforth termed topic concreteness and vehicle concreteness), extracted from the original studies. The availability of these measures varies, with some present for all items (100%) and others available for different subsets (down to 14%). For a portion of items in the module (124, 26.72%) we provide also a list of preferred interpretations, namely the features generated by participants to explain each metaphorical expression. To ensure consistency, original rating measures were standardized to homogeneous scales, while corpus-based measures were recalculated on up-to-date and open corpus resources. Additionally, new inclusiveness ratings were collected for all items.

*Collection of metaphors and ratings*

The metaphors and the relative psycholinguistic variables were drawn from studies that addressed figurative language processing with various methodologies and included a stage devoted to constructing and rating the stimuli. Additional information for each study is available in the Zenodo repository[63] and in each individual downloadable dataset. All studies were conducted on samples of native speakers of Italian, for a total of 630 undergraduate and graduate students. Across studies, participants were young adults with university education (316F; age: $M = 25.57$, $SD = 3.76$; education in years: $M = 16.52$, $SD = 2.50$). Data acquisition was conducted in compliance with the General Data Protection Regulation (GDPR) and following the guidelines of the Declaration of Helsinki. Data reuse in aggregated form was allowed in full compliance with the GDPR.

Forty-two nominal predicative metaphorical sentences, along with their matched literal counterparts, were taken from the study by Bambini et al. (2013)[64], which investigated reaction times during a sensicality judgment task in response to metaphors, metonymies, and approximations vs. literal and anomalous statements. The 42 metaphors appeared in the form *Quegli X sono Y* (Eng. Tr.: "Those Xs are Ys"), with X and Y being common nouns, e.g., *Quegli avvocati sono squali* (Eng. Tr.:

"Those lawyers are sharks"). Literal counterparts were obtained by replacing the topic with semantically compatible terms, e.g., *Quei pesci sono squali* (Eng. Tr.: "Those fish are sharks"). All items were rated for meaningfulness, familiarity, and difficulty by a sample of 85 native speakers of Italian (42F; age: $M = 26.85$, $SD = 3.80$; education in years: $M = 18.02$, $SD = 2.04$). Additionally, the same sample also provided cloze probability (CP) values for all sentences truncated before the target words, such as *Quegli X sono…* (Eng. Tr.: "Those Xs are…").

Sixty-four nominal predicative metaphorical sentences, along with their matched literal counterparts, were taken from the study by Bambini et al. (2016)[24], which analyzed the brain correlates of metaphor processing using the electroencephalography (EEG) technique. This study used stimuli constructed by expanding the set used in a previous neuroimaging study on metaphor comprehension Bambini et al. (2011)[49] and included metaphors in different sentential structures, to modulate the contextual information given across two experiments. In the first experiment, metaphors were embedded in a minimal context in the form *Sai che cos'è quell'X? È un Y* (Eng. Tr.: "Do you know what that X is? It's a Y"), with X and Y being common nouns, e.g., *Sai che cos'è quel soldato? È un leone* (Eng. Tr.: "Do you know what that soldier is? He's a lion"). In the second experiment, metaphors were embedded in a supportive context in the form *Quell' X è molto Z. È un Y* (Eng. Tr.: "That X is very Z. It's a Y"), with Z being an adjective that denoted a property linking X to Y, e.g., *Quel soldato è molto coraggioso. È un leone* (Eng. Tr.: "That soldier is very brave. He's a lion"). Literal counterparts were obtained by replacing the topic with a term in a literal relationship with the vehicle, e.g., *Sai che cos'è quel felino? È un leone* (Eng. Tr.: "Do you know what that feline is? It's a lion") and *Quel felino è molto coraggioso. È un leone* (Eng. Tr.: "That feline is very brave. It's a lion") respectively. CP values were collected from two groups of native speakers of Italian for sentences truncated before the target word: 15 participants for the minimal context sentences in the form *Quell' X è un…* (Eng. Tr.: "That X is a…"), and 14 for the supportive context sentences in the form *Quell'X è molto Z. È un…* ("That X is really Z. It's a…"). Additionally, the lexical frequency of the topic and vehicle was extracted from the CoLFIS database[65].

Eighty-two nominal predicative metaphorical sentences, along with their matched literal counterparts, were taken from the magnetoencephalography (MEG) study by Lago et al. (2024)[66]. The set overlapped significantly (62%) with the stimuli used in the study by Bambini et al. (2016)[24]. All sentences appeared in the form *Quell'X è un Y* (Eng. Tr.: "That X is a Y"), with X and Y being common nouns, e.g., *Quel matrimonio è una quercia* (Eng. Tr.: "That marriage is an oak"). Literal counterparts were obtained by replacing the topic with a term in a literal relationship with the vehicle, e.g., *Quell'albero è una quercia* (Eng. Tr.: "That tree is an oak"). All items were rated for familiarity by 39 native speakers of Italian (20F; age: $M = 27.05$, $SD = 4.54$, range = 20-43; education in years: $M = 16.69$, $SD = 2.44$, range = 11-21). Additionally, a sample of 17 native speakers of Italian (12F; age: $M = 29.00$, $SD = 6.29$, range = 22-46; education in years: $M = 16.00$, $SD = 2.74$, range = 13-21) provided CP and entropy values for all sentences truncated before the target words, such as *Quell' X è un…* (Eng. Tr.: "That X is a…"). Vehicle frequency was extracted from the itWAC corpus[67]; semantic distance between topic and vehicle was calculated using WEISS (Word-Embeddings Italian Semantic Space[68]).

One hundred and twenty-four nominal predicative metaphorical sentences formed the set used in the study by Canal et al. (2022)[27] to investigate the role of Theory of Mind (ToM) in processing physical vs. mental metaphors with EEG methods. All sentences appeared in the form *Spec Xs sono Ys* (Eng. Tr.: "Spec Xs are Ys"), with Spec being *certi/certe/alcuni/alcune/quelli/quelle* (Eng. Tr.: "certain/some/those") or the plural definite articles *i/gli/le* (Eng. Tr.: "the"), Xs being common nouns denoting human beings, Ys being common nouns denoting concrete non-human entities, and the relationship between X and Y being based either on physical characteristics, e.g., *Certi cantanti sono usignoli* (Eng. Tr.: "Some singers are nightingales") or mental ones, e.g., *Alcuni scolari sono uragani* (Eng. Tr.: "Some pupils are hurricanes"). No literal sentences were associated with the metaphorical ones in the original study. However, literal counterparts matched to 65 of the metaphors in Canal et al. (2022) were created for other EEG studies (IUSS NEPLab MetaImagery study and IUSS NEPLab MetaStep study) and included here. Metaphorical sentences were rated for familiarity, physicality, mentality, and aptness by 53 native speakers of Italian (40F; age: $M = 23.91$, range: 21–32; education in years: $M = 15.83$, range: 13–18). Vehicle frequency values were extracted from the CoLFIS database[65], while vehicle concreteness was sourced using the norms from Brysbaert et al. (2014)[69] after translation of items into English. Semantic distance between the topic and the vehicle was computed using WEISS[68]. To complement these measures, we also added the preferred metaphor interpretations collected by Battaglini et al. (2025)[70] from a sample of 76 Italian-speaking undergraduate students (age: $M = 22.08$, $SD = 1.52$; education in years: $M = 15$, $SD = 0.76$): participants were asked to write the meaning of each metaphor in a booklet, and were encouraged to identify key characteristics (features) essential for understanding each metaphor[68].

One hundred and twenty-eight metaphorical word pairs, along with their matched literal counterparts, were taken from the study by Bambini et al. (2024)[71], which investigated the processing costs of multimodal metaphors compared to verbal ones using the EEG technique. In the verbal condition, nominal metaphors in word pairs were used, in the *X – Y* form, e.g., *linguaggio – ponte* (Eng. Tr.: "language – bridge"), with X denoting abstract entities for half of the items and concrete ones for the other half, and Y denoting concrete entities. Literal counterparts were created by replacing X with a word in a literal relation with Y, e.g., *fiume – ponte* (Eng. Tr.: "river – bridge"). In the multimodal condition, the X from verbal pairs was combined with a picture representing Y, e.g., the image of a bridge. In the *Figurative Archive,* only verbal items are included. All items were rated for familiarity, difficulty, imageability, metaphoricity, number of interpretations, and strength of metaphorical interpretations by various subsamples from a pool of 122 native speakers of Italian (68F, age: $M = 24.34$, $SD = 1.97$). Vehicle frequency was extracted from the CoLFIS database[65]. Semantic distance between the two terms in each metaphorical pair was computed using WEISS[68].

Sixty predicate metaphors were taken from Frau et al. (2025)[62], which inquired into motor cortex involvement in action-language processing in two motor neuron diseases, Amyotrophic Lateral Sclerosis (ALS) and the *SPG4* variant of Hereditary Spastic Paraplegia (HSP-SPG4). The metaphors appeared in the form *Subj V (Ind)Obj*, with V being the vehicle expressed by a verb and (Ind)Obj being the topic expressed by a direct or indirect object[72], e.g., *Alice disegna il suo futuro con Alberto* (Eng. Tr.: "Alice is shaping her future with Alberto") and *Lisa corre verso l'amore con ingenuità* (Eng. Tr.: "Lisa rushes into love with innocence") respectively. Half of the sentences (30) described upper-limb-related action, as seen in the first example above, while the other half depicted lower-limb-related action, as in the second example above. Literal sentences were created by replacing the topic with an object in a literal relationship with the vehicle, e.g., *Il figlio disegna un ritratto della mamma* (Eng. Tr.: "The son draws a portrait of his mum"), *Francesca corre verso casa con il cane* (Eng. Tr.: "Francesca runs home with the dog"). All items were rated for meaningfulness and familiarity by a sample of 60 native speakers of Italian (35F; age: $M = 26.65$, $SD = 3.85$; education in years: $M = 15.80$, $SD = 2.15$).

Sixty-four nominal predicative metaphorical sentences, along with their matched literal counterparts, were taken from the unpublished IUSS NEPLab MetaBody study. Sentences appeared in the form: *Quel(quegli) X è(sono)* [*un*] *Y* (Eng. Tr.: "That(those) X(s) is(are) [a] Y(s)"). Xs and Ys were common nouns, with Xs referring to body parts, e.g., *Quei bicipiti sono sassi* (Eng. Tr.: "Those biceps are stones"), or to objects, e.g., *Quella casa è un gioiello* (Eng. Tr.: "That house is a jewel"). Literal counterparts were created by replacing the vehicle with a semantically compatible adjectival phrase: for the body-related items, e.g., *Quei bicipiti sono allenati* (Eng. Tr.: "Those biceps are trained"), and for the object-related items, e.g., *Quella casa è molto spaziosa* (Eng. Tr.: "That house is very spacious"). All items were rated for meaningfulness, familiarity, and body relatedness by 49 native speakers of Italian (27F; age: $M = 27.35$, $SD = 3.55$; education in years: $M = 15.82$, $SD = 2.76$). Vehicle frequency was extracted from the CoLFIS database[65].

Eighty nominal predicative metaphorical sentences, along with their matched literal counterparts, were taken from the unpublished IUSS NEPLab MetaEducation study. Of these, 42 were adapted from Bambini et al. (2013)[64], while 38 were newly created. Sentences were presented in the form *Quel(quegli) X è(sono)* [*un*] *Y* (Eng. Tr.: "That(those) X(s) is(are) [a] Y(s)"). Xs and Ys were common nouns, with Xs being either abstract or concrete topics. Each metaphor was embedded within a single-sentence context, e.g., *Nei momenti difficili le speranze sono stelle che illuminano l'anima* (Eng. Tr.: "In hard times hopes are stars that light up the soul"). Literal counterparts were created by modifying the topic of the metaphor and the context to ensure a literal interpretation, e.g., *Quelle luci nel cielo notturno sono stelle di galassie lontane* (Eng. Tr.: "Those lights in the night sky are stars of distant galaxies"). The items from Bambini et al. (2013)[64] were already rated for meaningfulness, familiarity, and difficulty. The newly created items were rated for the same measures by 49 native speakers of Italian (age: $M = 21.69$; $SD = 1.38$).

Moreover, for forty-two nominal predicative metaphors pooled from various studies above, values of imageability and physicality were added, taking them from the unpublished IUSS NEPLab MetaImagery study, which examined the role of visual mental imagery in metaphor processing using the EEG technique. This study used metaphors already included in the *Everyday Metaphors* module from other studies (i.e., Bambini et al. 2013, 2016; Canal et al. 2022[24,27,64], and the IUSS NEPLab MetaBody study) and collected additional ratings values of imageability and physicality from 64 native speakers of Italian (41F; age, $M = 24.13$, $SD = 2.47$; education in years, $M = 15.77$, $SD = 2.22$).

Overall, a total of 622 metaphors, with rating values for different measures, were extracted from nine studies. After removing duplicates, i.e., metaphors that appeared in more than one study in the same or a slightly different form (approximately 25% of all items), the *Everyday Metaphors* module of the *Figurative Archive* comprises 464 unique metaphors. Of these, 321 metaphors (69.18%) have a nominal predicative structure, e.g., *Quegli avvocati sono squali* (Eng. Tr.: "Those lawyers are sharks"), 60 (12.93%) are predicate metaphors, e.g., *Alice disegna il suo futuro con Alberto* (Eng.Tr.: "Alice is shaping her future with Alberto"), and 83 metaphors (17.89%) are nominal word pairs, e.g., *linguaggio – ponte* (Eng. Tr.: "language – bridge"). The 464 metaphors are displayed in the *Everyday Metaphors* module keeping their original structure of nominal word pairs, predicate metaphors, or nominal metaphors, with the latter type limited to the "X(s) is/are Y(s)", after dropping the broader context in the case of the study by Bambini et al. 2016 and in the IUSS NEPLab MetaEducation study), also reporting topics and vehicles in specific columns. The overall distribution of types in the *Archive* reflects the metaphor structures mostly used in psycho-neurolinguistic studies[35,37], being based mostly on nouns (87.07%) and partly on verbs (12.93%). Furthermore, the distribution of types closely aligns with the occurrence of metaphors in real-life language, where nominal and verbal metaphors dominate (63% of metaphorical occurrences in classroom discourse[73] and 80% of metaphors collected in corpus-based research[74]).

Since different studies collected different rating and corpus-based measures, some measures are more heavily represented than others (see Figure 1, lollipop plot on the left). Overall, the distribution of values for each dimension exhibits sufficient variability between items and highlights distinct characteristics of the stimuli across the dataset (Figure 1, density plots on the right). For instance, familiarity approximates a normal distribution, with most items showing moderate values. Conversely, body relatedness – defined as the inclusion of body parts or motor aspects in a sentence – shows a bimodal distribution. This may be because this dimension is represented only in one study (IUSS NEPLab MetaBody study), where items were constructed to be either body-related, thus scoring high in body relatedness, or object-related, thus scoring low in body relatedness. Mentality (i.e., how much a metaphor describes psychological qualities of the topic) also showed a bimodal distribution, while physicality (i.e., how much a metaphor describes physical qualities of the topic) closely resembled a normal distribution. This pattern seems to suggest that all metaphors can, to some extent, be interpreted physically, while a mental interpretation seems to be more specific for some metaphors (i.e., in our case, those originally constructed to express mental properties[26,27,70]. Regarding single-word measures, vehicles tended to be concrete across the dataset and, at the same time, topics displayed a broader range of concreteness values, aligning with the idea that metaphors often use more concrete, immediate terms to describe more abstract concepts (see Kövecses 2000 on emotion metaphors[75]) and offering the opportunity to test multimodal aspects of metaphor processing.

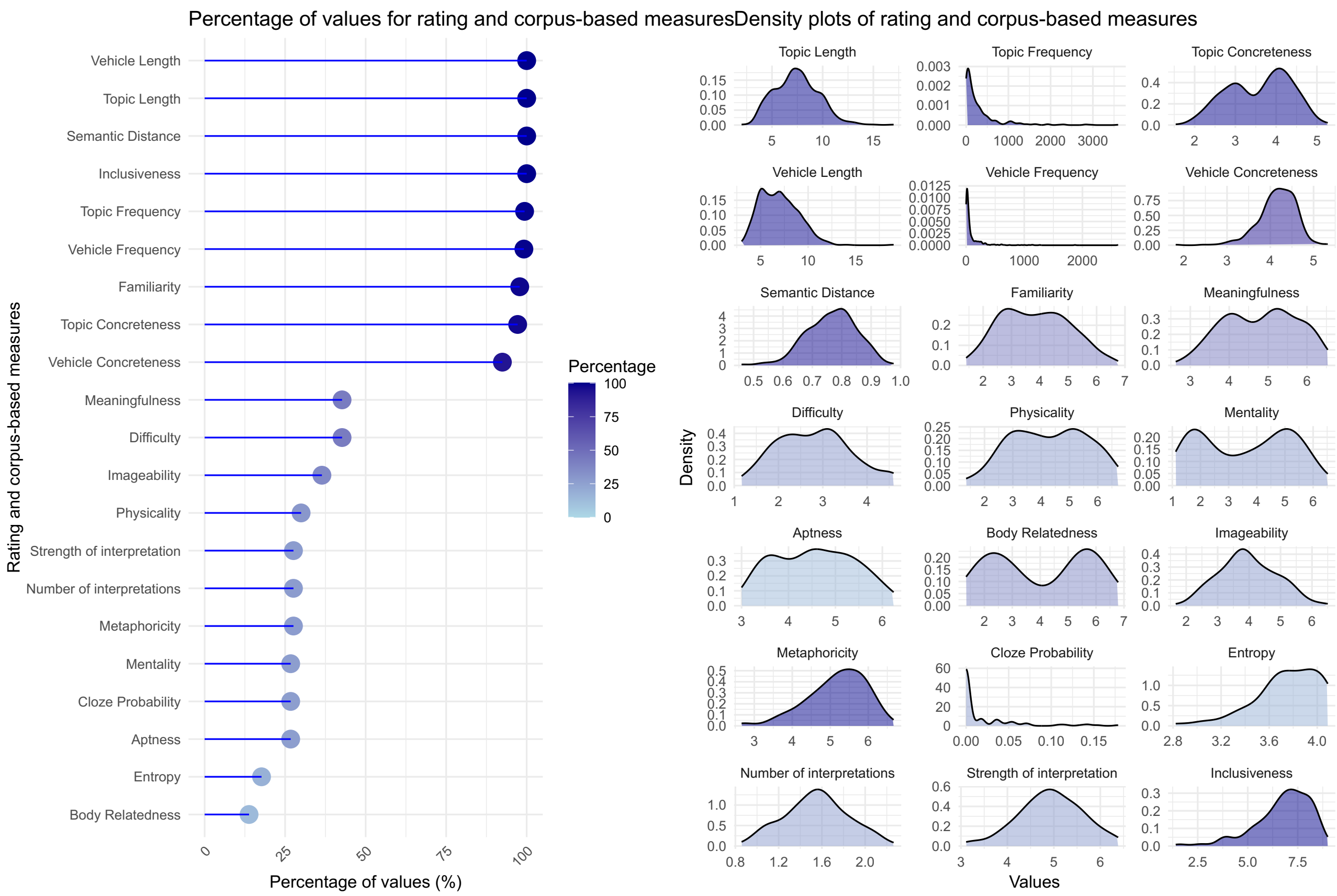

**Figure 1. Percentage of values for rating and corpus-based measures in the *Everyday Metaphors* module.** The lollipop plot on the left displays, for each variable, the percentages of metaphors that are described by that variable, over the total of the 464 metaphors from the *Everyday Metaphors* module. The density plots on the right illustrate the distribution of values for each variable. Darker shading indicates a higher proportion of items (i.e., percentage over the total) in the *Everyday Metaphors* module (lighter for lower coverage, e.g., body relatedness = 13.79%; darker for higher coverage, e.g., topic and vehicle length, semantic distance, and inclusiveness = 100%).

*Standardization and recalculation*

To ensure uniformity and reproducibility, some rating and corpus-based measures were recalculated or automatically re-extracted for the final dataset of 464 metaphors. Rating values, originally collected on either a 1-5 or 1-7 Likert scale, were first rescaled to a 1-7 Likert scale, the one most commonly employed across our studies. In the case of a metaphor included in more than one study and with multiple ratings for a given dimension, we averaged the available ratings for that dimension, considering that the samples of raters were homogeneous in terms of age and education (see the additional information for each study available in the dedicated files on Zenodo[63]). Before averaging, the consistency of the conceptualization of the given dimension across studies was verified: for example, while labeled differently (e.g., meaningfulness and sensicality), some dimensions were defined in the same way, and rating instructions provided to raters were equivalent (see the *Wiki* section of individual datasets available on Zenodo[63]). Corpus-based measures, including word frequency, word concreteness, and semantic distance, were extracted *de novo* for each metaphorical item in the *Archive*, using exclusively open-access tools for Italian. For example, absolute frequencies of topics and vehicles were extracted from the CoLFIS database[65], while vehicle concreteness and topic concreteness values were sourced from the MEGAHR-Crossling multilanguage dataset[76]. Semantic distance between the topic and vehicle was calculated using the Italian word embeddings from fastText[77], a set of pre-trained word vectors based on Common Crawl and Wikipedia. The main dataset of the *Figurative Archive* provides access to these recalculated and re-extracted values, while the original values are preserved in the downloadable version of each dataset for individual studies.

*Additional de novo ratings*

To assess the alignment of metaphors with current perspectives on inclusive language, ratings for inclusiveness were collected *de novo* for all items in the *Everyday Metaphors* module of the *Figurative Archive*. Prior research in the field of psycholinguistics[78,79] considered the related construct of offensiveness, which refers to those expressions that are perceived as having a negative impact on the sense of self and well-being of the interlocutor[71], often including a moral judgment[80]. For the purposes of the *Figurative Archive*, however, we chose to focus on the broader measure of inclusiveness, which we defined as a form of communication that recognizes diversity, conveys respect for others, is sensitive to differences, and promotes equal opportunities, based on the guidelines of the Linguistic Society of America (https://www.lsadc.org/guidelines_for_inclusive_language). In doing so, we accounted for a wider spectrum of potentially inappropriate uses of language, considering, in addition to expressions carrying a negative attitude toward certain social groups (such as the idea that wives are annoying, as a possible interpretation of the metaphor *Certe mogli sono martelli*, Eng. Tr.: "Some wives are hammers")[81,82], also expressions perpetuating positive stereotypes (such as the idea that girls are fragile, as a possible interpretation of the metaphor *Alcune fanciulle sono porcellane*, Eng. Tr.: "Some girls are porcelaines")[83].

To collect the rating, we developed a novel online questionnaire (hosted on LimeSurvey®), involving 15 Italian native speakers with experience in the study of language and ethical matters (graduate students and postgraduate fellows with backgrounds in linguistics, philosophy, and psychology; 9F; age: range = 18-34; education in years: range = 18-21). Participants were asked to rate each metaphor on a Likert scale, evaluating how respectful the metaphor was toward individual differences and how free it was from stereotypes and prejudices. Following previous studies on the related construct of offensiveness[84,85], we used a 9-point Likert scale, with lower ratings reflecting stronger stereotypical meanings and higher ratings indicating greater respectfulness. Metaphors were divided into three lists. The study was approved by the Ethics Committee of the Department of Brain and Behavioral Sciences of the University of Pavia (protocol number 123/2023). All participants provided written and informed consent, in accordance with the principles of the Declaration of Helsinki.

***Literary Metaphors***

The *Literary Metaphors* module includes 532 unique genitive metaphorical expressions in Italian sourced from literary works (poetry or prose), assembled from two studies conducted by members of the NEPLab. All metaphorical expressions appear in the form *X di Y* (Eng. Tr.: "X of Y"). A unique alphanumeric ID was assigned to each metaphorical expression based on the chronological order of the studies. Literal English translations are given for all metaphorical items, maintaining the terms as similar as possible to the Italian original. In addition to the author and the textual source from which they were extracted, each metaphor is accompanied by a set of relevant measures, obtained through rating tasks (meaningfulness, familiarity, difficulty, cloze probability, metaphor concreteness) and corpus-based indexes (frequency, topic and vehicle concreteness, readability index, semantic distance between the topic and vehicle). The availability of these measures varies, with some present for all items (100%) and others available for different subsets (down to 12%). Additional information for each study is available in the Zenodo repository[63] and in each individual downloadable dataset, detailed in the *Wiki* section of the available spreadsheets.

One hundred and fifteen genitive metaphors were taken from the study by Bambini et al. (2014)[42], which provided the first collection of Italian literary metaphors, half from poetry and half from prose, with psycholinguistic measures. The metaphorical expressions appeared in the form *X di Y* (Eng. Tr.: "X of Y"), with X and Y being common nouns. Of these, 24 (20.87%) expressions displayed the topic-vehicle (TV) order, with X being the topic and Y being the vehicle, e.g., *Labbra di rubino* (Eng. Tr.: "Lips of ruby"), and 91 (79.13%) displayed the vehicle-topic (VT) order, with X being the vehicle and Y being the topic, e.g., *Finestra dell'anima* (Eng. Tr.: "Window of the soul"). All items were rated in isolation (out of the literary context) for familiarity, metaphor concreteness, difficulty, and meaningfulness by 105 Italian native speakers (83F; age: $M$ = 23.00, $SD$ = 4.31). CP values were collected by truncating the metaphor after the preposition *di* (Eng. Tr.: "of"). A subset of 65 items was also rated for the same variables in the original context (average text length = 50 words) by 180 native speakers of Italian (145F; age: $M$ = 20.00, $SD$ = 2.50). Word frequency of the topic and vehicle was extracted from the CoLFIS database[65], phrase frequency was calculated in the Google search engine, and readability was measured through the Gulpease index[86].

Additionally, 417 genitive metaphors, 41% extracted from poetry and 59% extracted from prose, were taken from the unpublished IUSS NEPLab MetaLiterary study, which applied a semi-automatic methodology to extract metaphorical sentences from Italian prose and poetry literary texts. Initially, all occurrences of the *NOUN di NOUN* string (Eng. Tr.: "NOUN of NOUN") were isolated through PoS-tagging[87]. Following the approach outlined by Bambini et al. (2014)[42], expressions containing known metaphorical sources (such as physical locations and events) were manually reviewed. All extracted metaphorical expressions were in the form *X di Y* (Eng. Tr.: "X of Y"), with X and Y being common nouns. Of these, 118 (28.30%) expressions followed a topic-vehicle (TV) order, e.g., *Capelli di fiamma* (Eng. Tr.: "Hair of flame"), while 299 (71.70%) displayed the vehicle-topic (VT) order, e.g., *Nebbia di malinconia* (Eng. Tr.: "Fog of melancholy"). Lexical frequency of the topic and vehicle for each item was obtained from the CoLFIS database[65], and topic and vehicle concreteness values were sourced from the MEGAHR-Crossling multilanguage dataset[76]. Semantic distance between the topic and vehicle was calculated using the pre-trained Italian word embeddings from fastText[77].

Overall, a total of 532 metaphors were extracted from two studies and included in the *Literary Metaphors* module of the *Figurative Archive*, 390 (73.31%) with the VT order and 142 metaphors (26.69%) with the TV order.

Since different studies collected different rating and corpus-based measures, some measures are more heavily represented than others (see Figure 2, lollipop plot on the left). Overall, the distribution of values for each dimension exhibits sufficient variability between items, highlighting distinct characteristics of the stimuli across the dataset (Figure 2, density plots on the right).

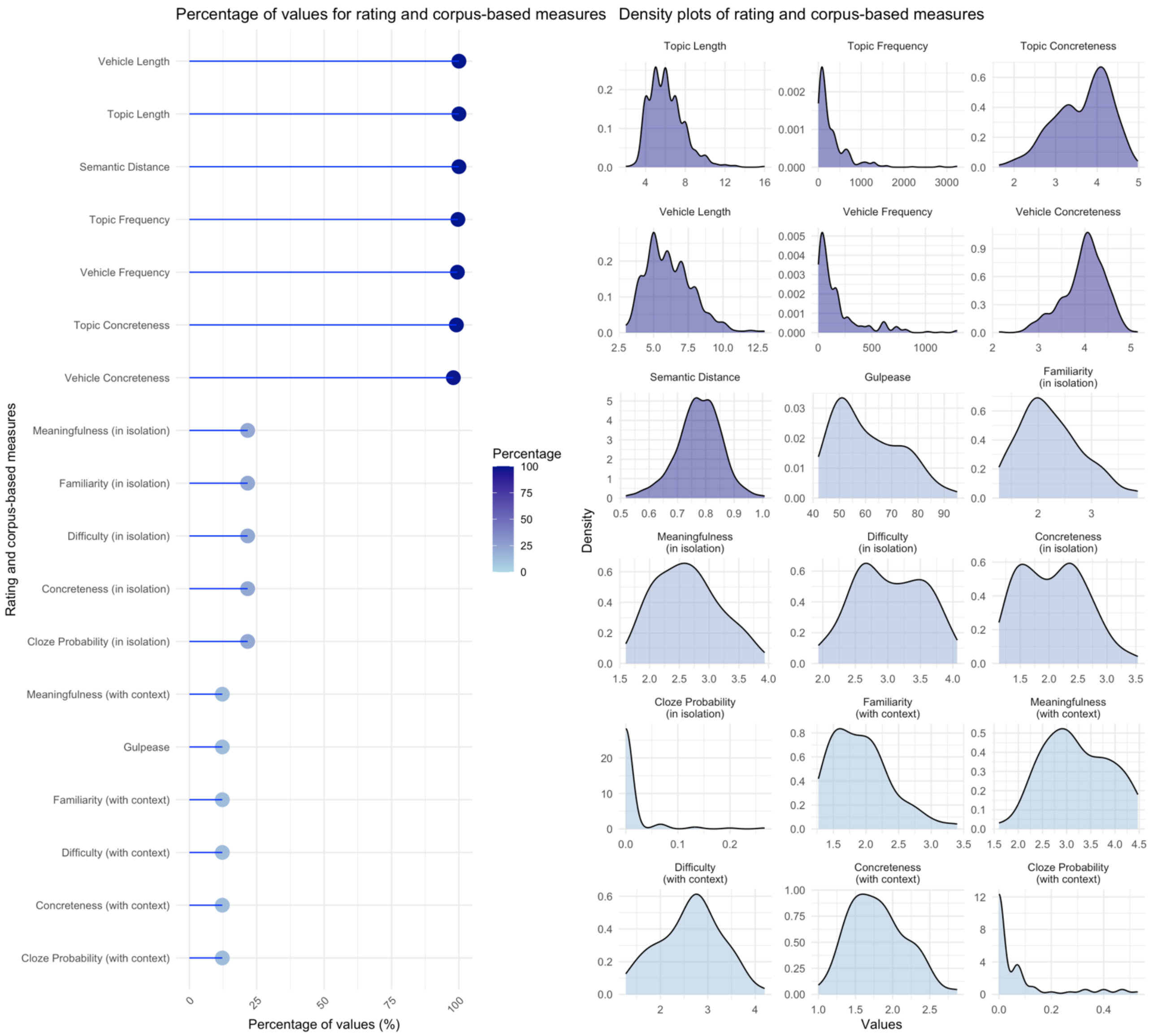

**Figure 2. Percentage values for rating and corpus-based measures in the *Literary Metaphors* module.** The lollipop plot on the left displays, for each variable, the percentages of metaphors that are described by that variable, over the total of the 532 metaphors from the *Literary Metaphors* module. The density plots on the right illustrate the distribution of values for each variable. Darker shading indicates a higher proportion of items (i.e., percentage over the total) in the *Literary Metaphors* module (lighter for lower coverage, e.g., cloze probability, metaphor concreteness, difficulty, familiarity, Gulpease index, and meaningfulness for metaphors presented with context = 12.22%; darker for higher coverage, e.g., topic and vehicle length and semantic distance = 100%).

To create a summary of the semantic domains covered in the *Literary Metaphors* module, we relied on the capability of LLMs to perform automatic topic modeling[88,89]. Methodologically, we used ChatGPT since it has been found to outperform human raters in topic detection[90]. Operationally, we prompted the model to cluster topics and vehicles into up to a feasible number (n = 10) of semantic classes, with a sufficient level of granularity[91], in line with previous works on literary texts[92]. Metaphorical topics and vehicles spanned a wide range of semantic classes (Figure 3): most topic words referred to natural elements (25.00%), e.g., *Cielo di perla* (Eng. Tr.: "Sky of pearl"), emotions or psychological states (15.22%), e.g., *Esplosione di dolore* (Eng. Tr.: "Explosion of pain"), and body and physical sensations (15.04%), e.g., *Viso di mela* (Eng. Tr.: "Face of apple"). Meanwhile, the automatic clustering revealed that the majority of vehicle words described natural elements (34.59%), e.g., *Fiume di lacrime* (Eng. Tr.: "River of tears"), material objects (25.38%), e.g., *Corpo di alabastro* (Eng. Tr.: "Body of alabaster"), or light and darkness (12.41%), e.g., *Lampo di gelosia* (Eng. Tr.: "Lightning of jealousy"). We hypothesize that the model's ability to assign clear semantic categories might have been enhanced by the metaphors being extracted via specific keywords (e.g., physical locations and events), thus limiting a priori the metaphors to specific conceptual domains and increasing classification reliability.

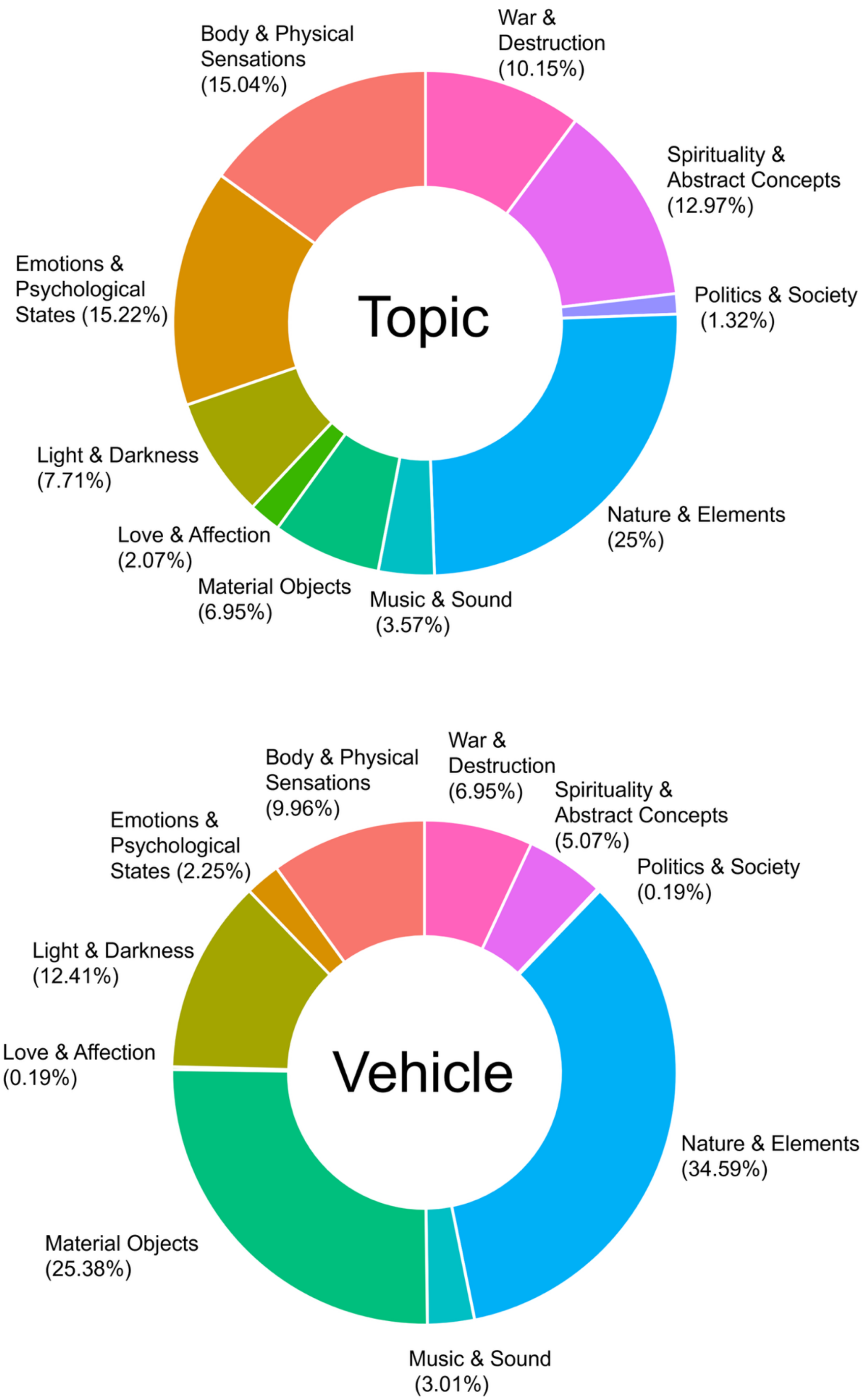

**Figure 3. Distribution of the ten semantic classes of metaphorical topics and vehicles in the *Literary Metaphors* module.** The upper panel displays the percentages for metaphor topics, while the lower panel the percentages for vehicles.

## Data Records

The *Figurative Archive* is available on Zenodo[63] at https://doi.org/10.5281/zenodo.14924803.
The materials in Zenodo consist of:

- The folder *Data*, including the main datasets for both the *Everyday Metaphors* and the *Literary Metaphors* modules (uploaded as .csv files).
- The folder *Original Files*, including the datasets of the individual studies (uploaded as .xlsx files). Each dataset contains metaphors and their ratings as originally collected in the published studies[24,27,62,64,71] and in the IUSS NEPLab MetaBody, MetaEducation, and MetaImagery. For the items from Canal et al. (2022)[27], a separate interoperable .xlsx file provides the preferred interpretations at the single-trial level[70]. For a subset of the items, the audio recordings, as used in the IUSS NEPLab MetaStep study are available (uploaded as .wav files) and included in a compressed subfolder.
- The folder *Code*, including the code for replicating the Technical Validation and the code for locally accessing the web interface.
- The folder *Tables*, including tables with additional information for each study.

The datasets in *Data* and *Original Files* are accompanied by a *Wiki* sheet that describes the content of their columns and provides additional information about the study, including detailed descriptive statistics about rating values and participants' demographics. Column headings are harmonized across datasets to ensure interoperability and facilitate comparisons of equivalent (or identical) measures. All materials are distributed under the Creative Commons Attribution 4.0 International (CC-BY) license.

## Technical validation

First, we assessed the reliability of the newly collected inclusiveness ratings using the Intraclass Correlation Coefficient (ICC), calculated across three item lists, each rated by five expert judges. The average ICC across lists was .58, indicating moderate inter-rater agreement (List A: ICC = .64, $F(215, 860) = 2.77$, $p < .001$; List B: ICC = .75, $F(216, 864) = 4.03$, $p < .001$; List C: ICC = .35, $F(30, 120) = 1.55$, $p = .052$). Rating values homogeneity was further evaluated using Cronbach's alpha. The average alpha was .99, indicating excellent internal consistency (List A: $\alpha = 1$; List B: $\alpha = 1$; List C: $\alpha = .97$), in line with previous ratings of metaphorical expressions[39,40,45]). Moreover, we tested for the difference in inclusiveness ratings between gender: the t-test came out inconclusive, with numerically higher rating values among males ($M = 7.53$, $SD = 1.95$) vs. females ($M = 6.98$, $SD = 1.53$), a difference that fell short of significance ($t(13) = -0.61$, $p = .554$).

Then, to validate the measures available for the two modules of the *Figurative Archive*, we conducted a series of correlations between all measures for each module, expecting patterns of association consistent with those reported in the literature. Based on the seminal work by Katz and colleagues[35], we expected everyday and literary metaphors to exhibit similar patterns. More specifically, for both modules, we anticipated a broad spectrum of robust correlations between classic rating measures, for instance, between metaphor familiarity and aptness, between familiarity and difficulty, and between difficulty and imageability[33,35,36]. Differently, we expected a more scattered pattern of associations between single-word corpus-based measures and rating ones, but significant associations between familiarity and metaphoricity on the one hand and features of the topic and the vehicle, such as word-level concreteness[31,93] and semantic distance between the two[94,95], on the other hand. Pearson's zero-order correlations were computed on data after standardization, recalculation and averaging of rating and corpus-based measures. To compensate for the high number of associations tested and to minimize Type I errors, *p*s were corrected with the False Discovery Rate (FDR) method by applying the Benjamini-Hochberg procedure.

***Everyday Metaphors***

Results generally confirmed our predictions. First, we found an extensive pattern of significant associations between most rating variables, as shown in Figure 4. Familiarity emerged as a key dimension, with very strong positive correlations with aptness ($r(122)=.92$) and meaningfulness ($r(196)=.85$), moderate correlations with imageability ($r(167)=.61$), strength of interpretation ($r(126) = .50$), and difficulty ($r(196)=-.42$), and weak correlations with mentality ($r(122)=.27$), number of interpretations ($r(126)=.32$), cloze probability ($r(112)=.30$), and metaphoricity ($r(126)=-.24$). These correlations align with patterns reported in the literature, confirming the large overlap between familiarity and aptness[33] and the moderate relation of familiarity with difficulty and imageability[37]. Moreover, difficulty positively correlated with metaphoricity ($r(124)=.43$) and was negatively related to imageability ($r(134)=-.69$), strength of interpretation ($r(125)=-.48$), and number of interpretations ($r(125)=-.37$). The latter two were also inter-related ($r(126)=.44$). Furthermore, imageability was positively associated with strength ($r(126)=.47$) and number of interpretations ($r(126)=.29$) and negatively related to metaphoricity ($r(126)=-.42$). Experimentally, these results align with established results reported in classical studies[35] and more recent work on literary metaphors[96].

Two other sets of significant correlations are worth noting. First, physicality was strongly and positively associated with imageability ($r(40)=.88$) and negatively associated with mentality ($r(122)=-.77$). This pattern is indicative of the relationship between metaphor and mental imagery processes[97–99]. Second, the novel measure of inclusiveness significantly correlated with body relatedness ($r(62)=-.59$), topic concreteness ($r(449)=-.24$), and difficulty ($r(196)=-.24$), suggesting that metaphors describing body parts might perpetuate stereotypical or offensive representations, besides being difficult (e.g., *Quelle labbra sono un canotto*, Eng. Tr.: "Those lips are a dinghy").

Concerning corpus-based measures, as expected, results showed a sparser pattern of correlations. Meaningful patterns of associations involved, in particular, word-level concreteness. Topic and vehicle concreteness were positively correlated with metaphor physicality ($r(130)=.26$ and $r(131)=.23$, respectively), while topic concreteness was negatively correlated with metaphor mentality ($r(115)=-.42$), as well as with metaphoricity ($r(124)=-.46$). These findings suggest that sensory-motor properties of the single terms and in particular of the topic directly impact the sensory-motor characterization of the whole figurative expression and possibly its metaphoricity. Interestingly, longer metaphorical topics (in characters) were related to greater metaphoricity values ($r(126)=.35$). Semantic distance stood out as the most relevant corpus-based measure: our analysis highlighted a positive association with metaphoricity ($r(126)=.27$) and negative relations with imageability ($r(167)=-.34$) and strength of interpretation ($r(126)=-.29$). These findings are indicative of the complexity of the semantic connections between topics and vehicles that underlie metaphorical relationships[35].

Overall, this correlation analysis supports the validity of the values reported in the *Everyday Metaphors* module, which can be used as an extensively normed set of experimental stimuli in the study of metaphor processing.

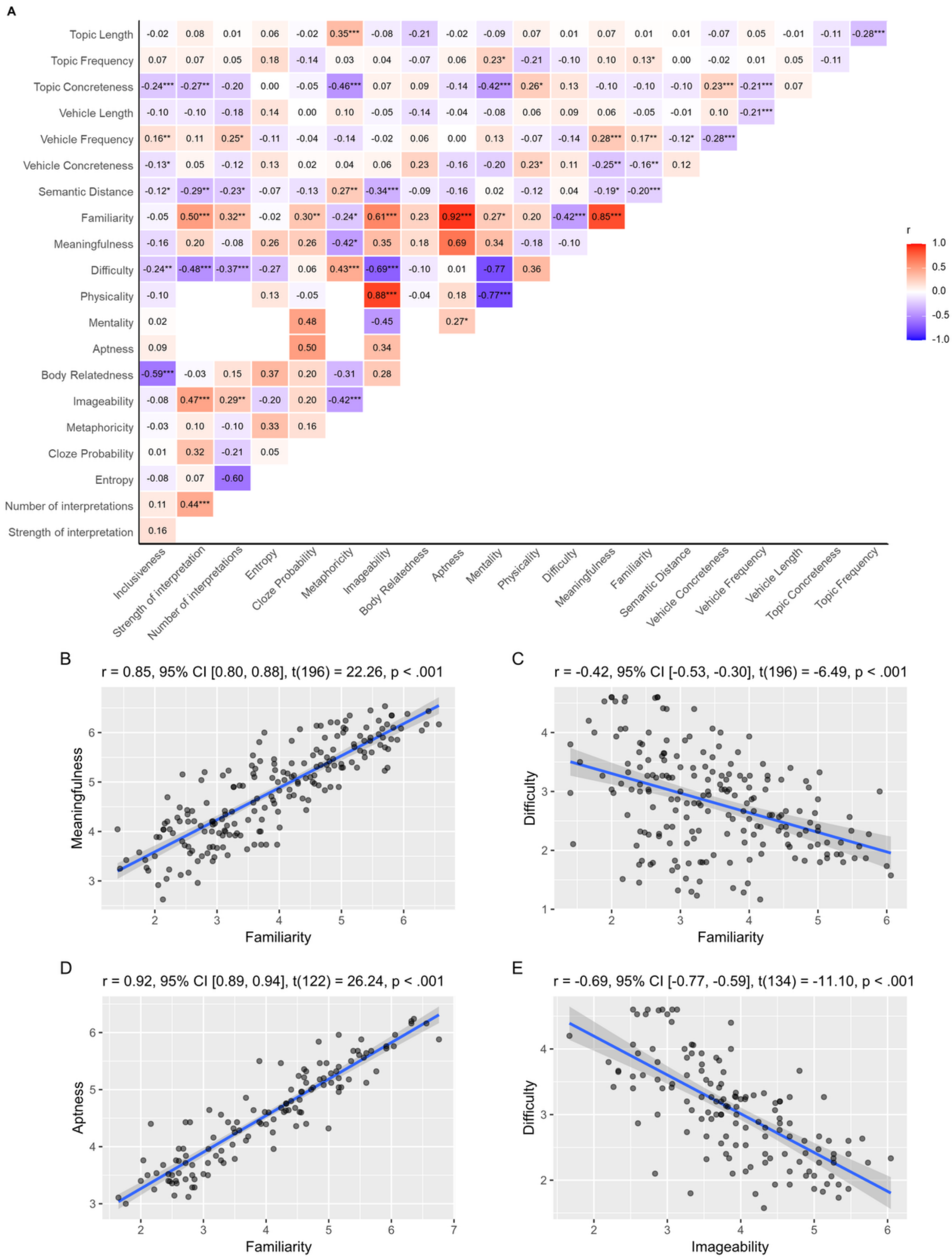

**Figure 4. Correlograms between rating and corpus-based measures of the *Everyday Metaphors* module.** Panel A shows the correlogram for all variables in the *Everyday Metaphors* module. The strength of the associations is represented by color (red for positive and blue for negative correlations), with significant (FDR-corrected) correlations marked by asterisks (**p*<.05, ***p*<.01, ****p*<.001). Panel B presents the scatterplot showing the relationship between familiarity and meaningfulness. Panel C illustrates the scatterplot showing the relationship between familiarity and difficulty. Panel D depicts the relationship between familiarity and aptness. Panel E shows the relationship between imageability and difficulty.

## *Literary Metaphors*

The correlational analysis confirmed our prediction (see Figure 5). Literary metaphors in isolation and those presented in context exhibited similar patterns, replicating the findings of Bambini and colleagues[42]. In both sets, familiarity emerged as a key dimension: it was moderately negatively correlated with difficulty (*r*(113)=-.59 for isolated metaphors; *r*(63)=-.37 for metaphors evaluated within context), and moderately positively correlated with metaphor concreteness (*r*(113)=.45 and *r*(63)=.64, respectively). Moreover, difficulty and meaningfulness were strongly negatively correlated (*r*(113)=-.88 and *r*(63)=-.73, respectively). In contrast, some associations emerged as specific to one set only. Notably, familiarity was moderately positively correlated with meaningfulness only for literary metaphors rated in isolation (*r*(113)=.67), while the association was negligible for metaphors embedded in context (*r*(63)=.08). Conversely, difficulty was moderately negatively correlated with metaphor concreteness only for literary metaphors evaluated within context (*r*(63)=-.43), an association not found for the set of metaphorical expressions rated in isolation (*r*(113)=-.20). Across the in-isolation and in-context sets, familiarity (*r*(63)=.56) and metaphor concreteness (*r*(63)=.69) showed moderate-to-strong positive correlations, indicating that these two dimensions are relatively stable for literary metaphors, regardless of contextual presentation condition.

With respect to corpus-based measures, the pattern of associations appeared more scattered. Topic and vehicle concreteness values at the word level stood out as key measures, in line with previous findings on everyday metaphors [31]: topic concreteness was negatively related to topic length (*r*(525)=-.29), with the latter also negatively associated with topic frequency (*r*(528)=-.34). Differently, vehicle concreteness was negatively related to vehicle frequency (*r*(519)=-.16), with the latter also negatively associated with vehicle length (*r*(527)=-.36). Moreover, topic concreteness was positively correlated with vehicle concreteness (*r*(514)=.26), and with concreteness of the whole metaphorical expression, both when rated in isolation (*r*(113)=.57) and in context (*r*(63)=.43). Another set of meaningful associations emerged between semantic distance and ratings collected for metaphors in isolation, specifically with metaphor concreteness (*r*(113)=-.30) and familiarity (*r*(113)=-.37), confirming the role of the semantic relationship between topic and vehicle in the generation of figurative meaning in literary items[100].

Overall, the correlation patterns between dimensions found in the *Everyday* module and those found in the *Literary* module closely resembled each other, in line with Katz et al.[35]. Rating measures clustered together, and so did corpus-based measures, with limited intercorrelation: this suggests that metaphor properties are weakly influenced by the lexical properties of topics and vehicles, with the notable exception of topic and vehicle concreteness, which impact especially metaphor concreteness and metaphoricity.

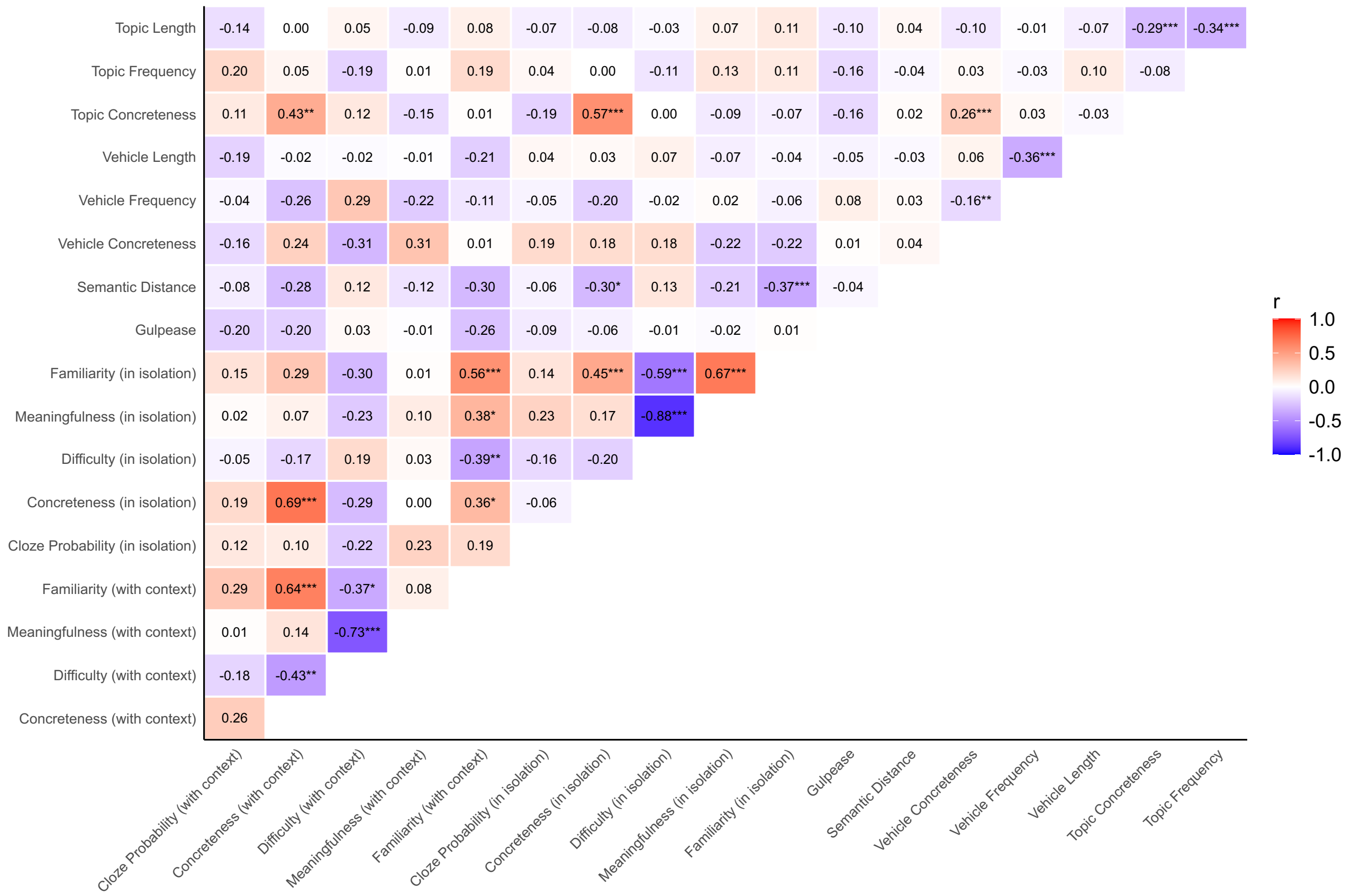

**Figure 5. Correlograms between rating and corpus-based measures of the *Literary Metaphors* module.** The strength of the associations is represented by color (red for positive and blue for negative correlations), with significant (FDR-corrected) correlations marked by asterisks (**p*<.05, ***p*<.01, ****p*<.001).

## Limitations and future directions

While the *Figurative Archive* is a valuable resource for advancing metaphor research, especially in Italian, some relevant aspects remain less covered and deserve attention for future expansions.

First, it would be relevant to address the variation of metaphor types, from nominal predicative to genitive and predicate metaphors. While the *Figurative Archive* includes a diverse range of metaphor types, encompassing the most widely attested types both in psycholinguistics[35,37] and in real-life uses of language[73,74], we acknowledge that a great variety exists across contexts, e.g., in advertising[71], and purposes, e.g., persuasive ones[101,102]. Future research might explore further the effects of using these metaphorical structures in other contexts.

Second, all rating measures are collected in samples of young and educated adults. Future studies should expand the present research by investigating the role of demographic factors such as age, which has been shown to have an effect on metaphor processing[28,51,52], and education, especially in relation to exposure to reading[103]. Given the modular architecture of the *Archive*, potential new datasets of metaphors, derived from future studies addressing these limitations, could be integrated into the resource, following the procedures indicated in the data-sharing policy available on Zenodo.

Third, offering English translations for our Italian metaphors inevitably confronts a main challenge in metaphor research, i.e., cross-linguistic and cross-cultural adaptation. Since metaphors are deeply embedded in cultural, cognitive, and linguistic contexts, their meanings often resist a direct translation[61,104]. The same vehicle may evoke different connotations[105], emotional resonances[106], or conceptual mappings across languages[107]. Thus, while the *Figurative Archive*, both in the web interface and in downloadable .csv files, offers a valuable entry point for identifying and comparing metaphorical patterns, it stands as the lower rung of a much larger ladder toward broader cross-linguistic metaphor research.

**Usage Notes**

The *Figurative Archive* is an initiative that aims to make available a large set of experimental stimuli, developed over the years for the study of metaphor processing, in a single resource. To pursue this aim, we standardized the data, originally collected from different participant samples and across various studies, by assigning a unique alphanumeric ID to each metaphor and ensuring uniformity in the labels for each rating and corpus-based measures. Metadata explaining each label is provided in the *Wiki* section of the web interface. Furthermore, rating measures were aggregated by rescaling to a 7-point Likert scale and averaging across studies where necessary. Corpus-based measures were uniformly re-collected, often using open-access tools to ensure reproducibility. The result of this process is a harmonized and cohesive archive of experimental stimuli that supports the reuse of existing materials, also for large-scale studies. Original data are still available for consultation to retrieve measures used in the individual studies.

Due to its modular nature, the *Figurative Archive* is well-suited for future expansions, both by the original team of contributors and by the broader academic community. Since it is an ongoing initiative, the participation of researchers in metaphor studies is welcomed and encouraged, promoting resource sharing and allowing broader replicability of results, in adherence to the FAIR principles (Findable, Accessible, Interoperable, and Reusable). Future expansions will undergo a process to guarantee their quality and, although each module can maintain its specificity, contributors will be asked to conform to a standardized basic data structure established in the dedicated data-sharing policy.

In addition to the Zenodo repository described in the Data Record section, to ensure a user-friendly experience with the *Figurative Archive,* we developed a web-based graphical user interface in R[108] with the *Shiny*[109] and *shinydashboard* packages[110]. The web interface is freely accessible at https://neplab.shinyapps.io/FigurativeArchive/ and follows the modular architecture of the *Figurative Archive*, currently comprising two main parts: the *Everyday Metaphors* module and the *Literary Metaphors* module. The web interface is particularly useful for accessing the standardized, re-scaled ratings for each measure, the averaged value for items with multiple entries across (rating) studies, and the re-extracted shared corpus-based measures from single resources, as well as interpretation data when available. In the web interface, the data and information of each module can be explored through four sections: *Wiki*, *Explore Dataset*, *Download*, and *References*. The *Wiki* section provides a comprehensive description of the dataset, including details about column headings and content. The *Explore Dataset* section, which constitutes the core of the *Figurative Archive,* allows users to browse the dataset, view each metaphor, and examine its rating and corpus-based measures, along with, when available, a summary (i.e., the three most frequent features) of its preferred interpretations (*Data* subsection – Figure 6A). The section also allows users to visualize data distribution through density plots, histograms, and rug plots (*Density Plot* subsection – Figure 6B) and to explore the relationships between variables through interactive scatterplots *(Scatter Plot* subsection – Figure 6C and 6D). In the *Download* section, users can download the datasets of the individual studies. Finally, the *Reference* section provides the complete list of references, including PDFs of open-access publications.

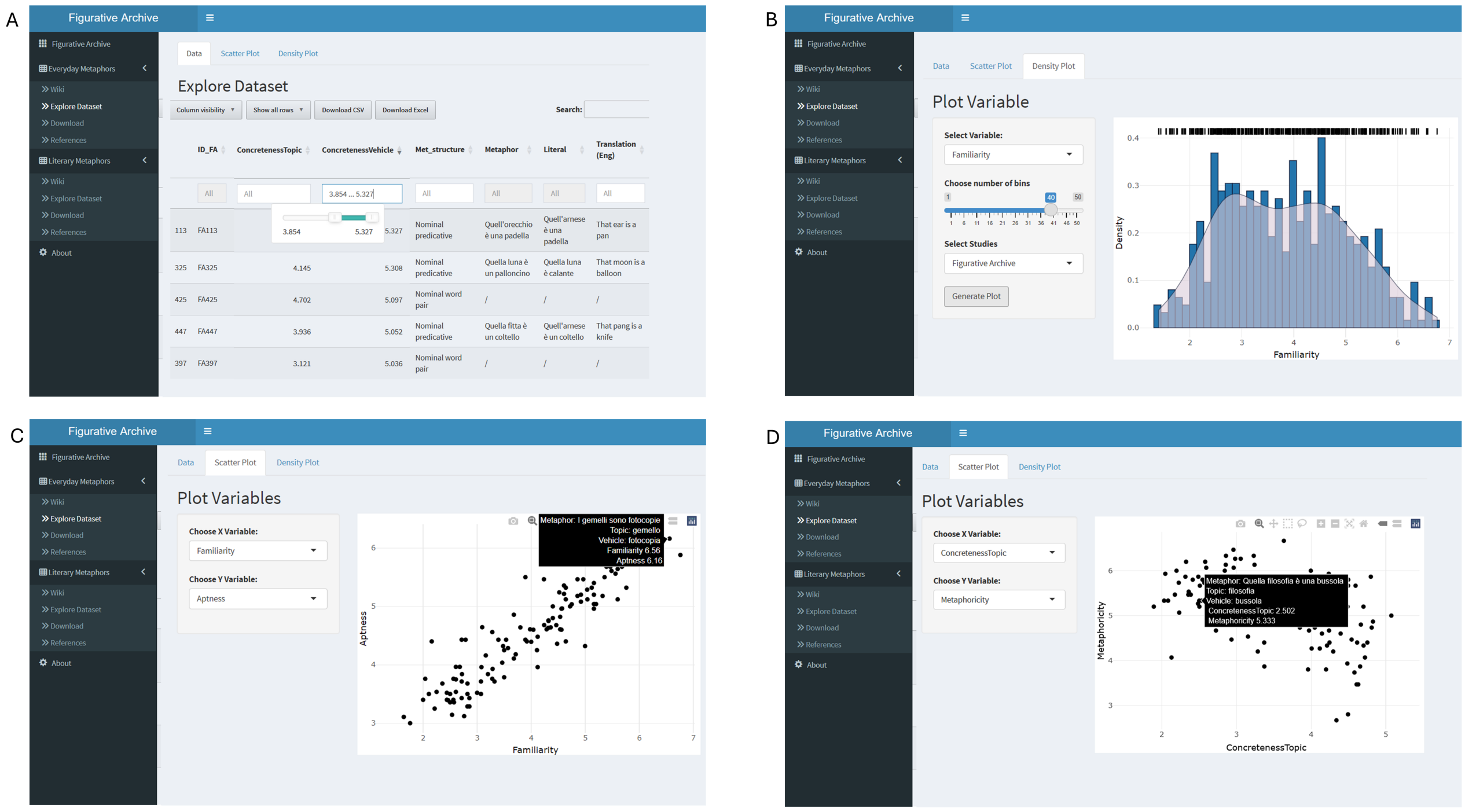

**Figure 6. Sections of the *Figurative Archive* web interface.** Panel A shows the *Data* subsection of *Explore Dataset,* featuring an example from the *Everyday Metaphors* module displaying a search filtered for specific values of vehicle concreteness. Panel B shows the *Density Plot* subsection of *Explore Dataset*, with histogram and density plot illustrating the distribution of familiarity ratings for metaphors from the *Everyday Metaphors* module. Panels C and D show the *Scatter Plot* subsection of *Explore Dataset*, with two different variable combinations plotted: familiarity and aptness in Panel C and topic concreteness and metaphoricity in Panel D. All panels show examples from the *Everyday Metaphors* module of the *Figurative Archive*, and the same structure applies to the *Literary Metaphors* module.

## Data Availability

The *Figurative Archive*, including the main datasets and the datasets of the original studies, is available on Zenodo[63] at https://doi.org/10.5281/zenodo.14924803.

## Code Availability

The codes to reproduce the Technical Validation and to locally access the interface are provided as R scripts in the Zenodo repository[63] at https://doi.org/10.5281/zenodo.14924803.

## Author Contribution (CRediT)

Maddalena Bressler: Methodology, Validation, Investigation, Formal analysis, Data Curation, Visualization, Writing - Original Draft
Veronica Mangiaterra: Methodology, Software, Validation, Investigation, Formal analysis, Data Curation, Visualization, Writing - Original Draft
Paolo Canal: Investigation, Data Curation, Writing - Review & Editing
Federico Frau: Investigation, Data Curation, Writing - Review & Editing
Fabrizio Luciani: Investigation, Writing - Review & Editing
Biagio Scalingi: Investigation, Writing - Review & Editing
Chiara Barattieri di San Pietro: Investigation, Writing - Review & Editing
Chiara Battaglini: Investigation, Writing - Review & Editing
Chiara Pompei: Investigation, Writing - Review & Editing
Fortunata Romeo: Investigation, Writing - Review & Editing
Luca Bischetti: Conceptualization, Methodology, Validation, Formal analysis, Visualization, Supervision, Writing - Original Draft

Valentina Bambini: Conceptualization, Methodology, Data Curation, Supervision, Writing - Original Draft, Resources, Funding acquisition

**Competing interests**
The authors declare no competing interests.

**Acknowledgements**
This work received support from the European Research Council under the EU's Horizon Europe programme, ERC Consolidator Grant "PROcessing MEtaphors: Neurochronometry, Acquisition and Decay, PROMENADE" [101045733]. The content of this article is the sole responsibility of the authors. The European Commission or its services cannot be held responsible for any use that may be made of the information it contains. We thank Prof. Filippo Domaneschi for his valuable input in refining the definition and conceptual scope of the inclusiveness measure. We are also grateful to Prof. Gloria Cappelli and Cordelia Katrina Bacci for their assistance in refining the English adaptation of metaphorical items based on verbs.

1. Holyoak, K. J. & Stamenković, D. Metaphor comprehension: A critical review of theories and evidence. *Psychol Bull* **144**, 641–671 (2018).
2. Rapp, A. M., Mutschler, D. E. & Erb, M. Where in the brain is nonliteral language? A coordinate-based meta-analysis of functional magnetic resonance imaging studies. *Neuroimage* **63**, 600–610 (2012).
3. Huang, Y., Huang, J., Li, L., Lin, T. & Zou, L. Neural network of metaphor comprehension: an ALE meta-analysis and MACM analysis. *Cerebral Cortex* **33**, 10918–10930 (2023).
4. Glucksberg, S. The psycholinguistics of metaphor. *Trends Cogn Sci* **7**, 92–96 (2003).
5. Yuan, G. & Sun, Y. A bibliometric study of metaphor research and its implications (2010–2020). *Southern African Linguistics and Applied Language Studies* **41**, 227–247 (2023).
6. Peng, Z. & Khatin-Zadeh, O. Research on metaphor processing during the past five decades: a bibliometric analysis. *Humanit Soc Sci Commun* **10**, 928 (2023).
7. Zhao, X., Zheng, Y. & Zhao, X. Global bibliometric analysis of conceptual metaphor research over the recent two decades. *Front Psychol* **14**, 1042121 (2023).
8. Coulson, S. Metaphor Comprehension and the Brain. in *The Cambridge Handbook of Metaphor and Thought* (ed. Gibbs, R. W. Jr.) 177–194 (Cambridge University Press, 2008).
9. *Methods in Cognitive Linguistics*. (John Benjamins Publishing Company, 2007). doi:10.1075/hcp.18.
10. Noveck, I. *Experimental Pragmatics: The Making of a Cognitive Science*. (Cambridge University Press, 2018).
11. Canal, P. & Bambini, V. Pragmatics Electrified. in *Language electrified. Principles, methods, and future perspectives of investigation* (eds. Grimaldi, M., Brattico, E. & Shtyrov, Y.) 583–612 (Humana, 2023). doi:10.1007/978-1-0716-3263-5_18.
12. Bambini, V. & Domaneschi, F. Twenty years of experimental pragmatics. New advances in scalar implicature and metaphor processing. *Cognition* **244**, 105708 (2024).
13. Bischetti, L., Frau, F. & Bambini, V. Neuropragmatics. in *The Handbook of Clinical Linguistics, Second Edition* (eds. Ball, M. J., Müller, N. & Spencer, E.) 41–54 (Wiley, 2024). doi:10.1002/9781119875949.ch4.
14. Cuccio, V. The figurative brain. in *The Routledge Handbook of Semiosis and the Brain* (eds. García, A. M. & Ibáñez, A.) 130–144 (Routledge, 2022). doi:10.4324/9781003051817-11.
15. Rossetti, I., Brambilla, P. & Papagno, C. Metaphor Comprehension in Schizophrenic Patients. *Front Psychol* **9**, 670 (2018).
16. Bambini, V. *et al.* A leopard cannot change its spots: A novel pragmatic account of concretism in schizophrenia. *Neuropsychologia* **139**, 107332 (2020).

17. Littlemore, J. & Low, G. Metaphoric Competence, Second Language Learning, and Communicative Language Ability. *Appl Linguist* **27**, 268–294 (2006).
18. Werkmann Horvat, A., Bolognesi, M., Littlemore, J. & Barnden, J. Comprehension of different types of novel metaphors in monolinguals and multilinguals. *Lang Cogn* **14**, 401–436 (2022).
19. Jacobs, A. M. (Neuro-)Cognitive poetics and computational stylistics. *Sci Study Lit* **8**, 165–208 (2018).
20. Jacobs, A. M. Neurocognitive poetics: methods and models for investigating the neuronal and cognitive-affective bases of literature reception. *Front Hum Neurosci* **9**, 186 (2015).
21. Bolognesi, M. & Werkmann Horvat, A. *The Metaphor Compass*. (Routledge, 2022). doi:10.4324/9781003041221.
22. Lai, V. T., Curran, T. & Menn, L. Comprehending conventional and novel metaphors: An ERP study. *Brain Res* **1284**, 145–155 (2009).
23. Schmidt, G. L. & Seger, C. A. Neural correlates of metaphor processing: The roles of figurativeness, familiarity and difficulty. *Brain Cogn* **71**, 375–386 (2009).
24. Bambini, V., Bertini, C., Schaeken, W., Stella, A. & Di Russo, F. Disentangling Metaphor from Context: An ERP Study. *Front Psychol* **7**, 559 (2016).
25. Yang, J. & Shu, H. Involvement of the Motor System in Comprehension of Non-Literal Action Language: A Meta-Analysis Study. *Brain Topogr* **29**, 94–107 (2016).
26. Lecce, S., Ronchi, L., Del Sette, P., Bischetti, L. & Bambini, V. Interpreting physical and mental metaphors: Is Theory of Mind associated with pragmatics in middle childhood? *J Child Lang* **46**, 393–407 (2019).
27. Canal, P. *et al.* N400 differences between physical and mental metaphors: The role of Theories of Mind. *Brain Cogn* **161**, 105879 (2022).
28. Ceccato, I. *et al.* Aging and the Division of Labor of Theory of Mind Skills in Metaphor Comprehension. *Top Cogn Sci* https://doi.org/10.1111/tops.12785 (2025) doi:10.1111/tops.12785.
29. McQuire, M., McCollum, L. & Chatterjee, A. Aptness and beauty in metaphor. *Lang Cogn* **9**, 316–331 (2017).
30. Jones, L. L. & Estes, Z. Roosters, robins, and alarm clocks: Aptness and conventionality in metaphor comprehension. *J Mem Lang* **55**, 18–32 (2006).
31. Al-Azary, H. & Buchanan, L. Novel metaphor comprehension: Semantic neighbourhood density interacts with concreteness. *Mem Cognit* **45**, 296–307 (2017).
32. Reid, N. J., Al-Azary, H. & Katz, A. N. Cognitive Factors Related to Metaphor Goodness in Poetic and Non-literary Metaphor. *Metaphor Symb* **38**, 130–148 (2023).
33. Thibodeau, P. H., Sikos, L. & Durgin, F. H. Are subjective ratings of metaphors a red herring? The big two dimensions of metaphoric sentences. *Behav Res Methods* **50**, 759–772 (2018).
34. Schmidt, G. L., Kranjec, A., Cardillo, E. R. & Chatterjee, A. Beyond Laterality: A Critical Assessment of Research on the Neural Basis of Metaphor. *Journal of the International Neuropsychological Society* **16**, 1–5 (2010).
35. Katz, A. N., Paivio, A., Marschark, M. & Clark, J. M. Norms for 204 Literary and 260 Nonliterary Metaphors on 10 Psychological Dimensions. *Metaphor and Symbolic Activity* **3**, 191–214 (1988).
36. Campbell, S. J. & Raney, G. E. A 25-year replication of Katz et al.'s (1988) metaphor norms. *Behav Res Methods* **48**, 330–340 (2016).
37. Cardillo, E. R., Schmidt, G. L., Kranjec, A. & Chatterjee, A. Stimulus design is an obstacle course: 560 matched literal and metaphorical sentences for testing neural hypotheses about metaphor. *Behav Res Methods* **42**, 651–664 (2010).

38. Cardillo, E. R., Watson, C. & Chatterjee, A. Stimulus needs are a moving target: 240 additional matched literal and metaphorical sentences for testing neural hypotheses about metaphor. *Behav Res Methods* **49**, 471–483 (2017).
39. Roncero, C. & de Almeida, R. G. Semantic properties, aptness, familiarity, conventionality, and interpretive diversity scores for 84 metaphors and similes. *Behav Res Methods* **47**, 800–812 (2015).
40. Citron, F. M. M., Lee, M. & Michaelis, N. Affective and psycholinguistic norms for German conceptual metaphors (COMETA). *Behav Res Methods* **52**, 1056–1072 (2020).
41. Müller, N., Nagels, A. & Kauschke, C. Metaphorical expressions originating from human senses: Psycholinguistic and affective norms for German metaphors for internal state terms (MIST database). *Behav Res Methods* **54**, 365–377 (2022).
42. Bambini, V., Resta, D. & Grimaldi, M. A Dataset of Metaphors from the Italian Literature: Exploring Psycholinguistic Variables and the Role of Context. *PLoS One* **9**, e105634 (2014).
43. Krennmayr, T. & Steen, G. VU Amsterdam Metaphor Corpus. in *Handbook of Linguistic Annotation* (eds. Ide, N. & Pustejovsky, J.) 1053–1071 (Springer Netherlands, 2017). doi:10.1007/978-94-024-0881-2_39.
44. Stamenković, D., Milenković, K. & Dinčić, J. Studija normiranja književnih i neknjiževnih metafora iz srpskog jezika [A norming study of Serbian literary and nonliterary metaphors]. *Zbornik Matice srpske za filologiju i lingvistiku* **62**, 89–104 (2019).
45. Huang, J., Chen, L., Huang, Y., Chen, Y. & Zou, L. COGMED: a database for Chinese olfactory and gustatory metaphor. *Humanit Soc Sci Commun* **11**, 1080 (2024).
46. Wang, X. Normed dataset for novel metaphors, novel similes, literal and anomalous sentences in Chinese. *Front Psychol* **13**, 922722 (2022).
47. Neumann, C. Is Metaphor Universal? Cross-Language Evidence From German and Japanese. *Metaphor Symb* **16**, 123–142 (2001).
48. Milenković, K., Tasić, M. & Stamenković, D. Influence of translation on perceived metaphor features: quality, aptness, metaphoricity, and familiarity. *Linguistics Vanguard* **10**, 285–296 (2024).
49. Bambini, V., Gentili, C., Ricciardi, E., Bertinetto, P. M. & Pietrini, P. Decomposing metaphor processing at the cognitive and neural level through functional magnetic resonance imaging. *Brain Res Bull* **86**, 203–216 (2011).
50. Poth, C. N. Fostering Equity and Diversity Through Essential Mixed Methods Research Inclusive Language Practices. *J Mix Methods Res* **18**, 110–114 (2024).
51. Bischetti, L. *et al.* Assessment of pragmatic abilities and cognitive substrates (APACS) brief remote: a novel tool for the rapid and tele-evaluation of pragmatic skills in Italian. *Lang Resour Eval* **58**, 951–979 (2024).
52. Bischetti, L. *et al.* Development and Validation of a Rapid Tool to Measure Pragmatic Abilities: The Brief Assessment of Pragmatic Abilities and Cognitive Substrates (APACS Brief). *Behavioral Sciences* **15**, 107 (2025).
53. Barattieri di San Pietro, C., Frau, F., Mangiaterra, V. & Bambini, V. The pragmatic profile of ChatGPT: Assessing the communicative skills of a conversational agent. *Sistemi intelligenti* 379–400 (2023).
54. Lombardi, A. & Lenci, A. Doing Things with Words: Rethinking Theory of Mind Simulation in Large Language Models. in *CLiC-it 2025: Eleventh Italian Conference on Computational Linguistics* (2025).
55. Mangiaterra, V., Barattieri Di San Pietro, C., Frau, F., Bambini, V. & Al-Azary, H. On choosing the vehicles of metaphors without a body: evidence from Large Language Models. in *Proceedings of the 2nd Workshop on Analogical Abstraction in Cognition, Perception, and Language (Analogy-Angle II)* 37–44 (Association for Computational Linguistics, Stroudsburg, PA, USA, 2025). doi:10.18653/v1/2025.analogyangle-1.4.

56. Pedinotti, P., Di Palma, E., Cerini, L. & Lenci, A. A howling success or a working sea? Testing what BERT knows about metaphors. in *Proceedings of the Fourth BlackboxNLP Workshop on Analyzing and Interpreting Neural Networks for NLP* 192–204 (Association for Computational Linguistics, Stroudsburg, PA, USA, 2021). doi:10.18653/v1/2021.blackboxnlp-1.13.
57. Ichien, N., Stamenković, D. & Holyoak, K. J. Large Language Model Displays Emergent Ability to Interpret Novel Literary Metaphors. *Metaphor Symb* **39**, 296–309 (2024).
58. Alice, S., Capone, L., Lebani, G. E. & Lenci, A. BAMBI: Developing BAby language Models for Italian. *Lingue e linguaggio* 83–102 (2025).
59. Attanasio, G. *et al.* CALAMITA: Challenge the Abilities of LAnguage Models in ITAlian. in *Proceedings of the 10th Italian Conference on Computational Linguistics (CLiC-it 2024)* (Pisa, Italy, 2024).
60. Blasi, D. E., Henrich, J., Adamou, E., Kemmerer, D. & Majid, A. Over-reliance on English hinders cognitive science. *Trends Cogn Sci* **26**, 1153–1170 (2022).
61. Kövecses, Z. *Metaphor in Culture*. (Cambridge University Press, 2005). doi:10.1017/CBO9780511614408.
62. Frau, F. *et al.* Testing motor grounding and somatotopic effects for literal and figurative action-language in Motor Neuron Diseases: a multiple-case analysis. Preprint at https://doi.org/10.31234/osf.io/mk9wh_v1 (2025).
63. IUSS Neurolinguistics and Experimental Pragmatics Laboratory (NEPLab). ERC_Cog PROMENADE - WP1: Figurative Archive. *Zenodo* https://doi.org/10.5281/zenodo.14924803 (2025) doi:10.5281/zenodo.14924803.
64. Bambini, V., Ghio, M., Moro, A. & Schumacher, P. B. Differentiating among pragmatic uses of words through timed sensicality judgments. *Front Psychol* **4**, 938 (2013).
65. Bertinetto, P. M. *et al.* Corpus e Lessico di Frequenza dell'Italiano Scritto (CoLFIS). (2005).
66. Lago, S., Zago, S., Bambini, V. & Arcara, G. Pre-Stimulus Activity of Left and Right TPJ in Linguistic Predictive Processing: A MEG Study. *Brain Sci* **14**, 1014 (2024).
67. Baroni, M., Bernardini, S., Ferraresi, A. & Zanchetta, E. The WaCky wide web: a collection of very large linguistically processed web-crawled corpora. *Lang Resour Eval* **43**, 209–226 (2009).
68. Marelli, M. Word-embeddings Italian semantic spaces: A semantic model for psycholinguistic research. *Psihologija* **50**, 503–520 (2017).
69. Brysbaert, M., Warriner, A. B. & Kuperman, V. Concreteness ratings for 40 thousand generally known English word lemmas. *Behav Res Methods* **46**, 904–911 (2014).
70. Battaglini, C. *et al.* Imagers and mentalizers: capturing individual variation in metaphor interpretation via intersubject representational dissimilarity. *Lingue e linguaggio* 61–81 (2025).
71. Bambini, V. *et al.* The costs of multimodal metaphors: comparing ERPs to figurative expressions in verbal and verbo-pictorial formats. *Discourse Process* **61**, 44–68 (2024).
72. Taverniers, M. Metaphor. in *Handbook of Pragmatics* 1–36 (John Benjamins Publishing Company, Amsterdam, 2003). doi:10.1075/hop.8.met3.
73. Cameron, L. *Metaphor in Educational Discourse*. (Continuum, 2003).
74. Berber Sardinha, T. Metaphor probabilities in corpora. in *Confronting Metaphor in Use: An applied linguistic approach* (eds. Zanotto, M. S., Cameron, L. & Cavalcanti, M. C.) 127–147 (John Benjamins Publishing Company, 2008). doi:10.1075/pbns.173.09ber.
75. Kövecses, Z. *Metaphor and Emotion. Language, Culture, and Body in Human Feeling*. (Cambridge University Press, Cambridge, 2000).
76. Ljubešić, N., Fišer, D. & Peti-Stantić, A. Predicting Concreteness and Imageability of Words Within and Across Languages via Word Embeddings. in *Proceedings of The Third Workshop on Representation Learning for NLP* 217–222 (Association for Computational Linguistics, Stroudsburg, PA, USA, 2018). doi:10.18653/v1/W18-3028.

77. Grave, E., Bojanowski, P., Gupta, P., Joulin, A. & Mikolov, T. Learning Word Vectors for 157 Languages. in *Proceedings of the Eleventh International Conference on Language Resources and Evaluation (LREC 2018)* (European Language Resources Association (ELRA), Miyazaki, Japan, 2018).
78. Janschewitz, K. Taboo, emotionally valenced, and emotionally neutral word norms. *Behav Res Methods* **40**, 1065–1074 (2008).
79. Sulpizio, S. *et al.* Taboo language across the globe: A multi-lab study. *Behav Res Methods* **56**, 3794–3813 (2024).
80. Culpeper, J. & Haugh, M. The metalinguistics of offence in (British) English. *Journal of Language Aggression and Conflict* **9**, 185–214 (2021).
81. Cepollaro, B., Sulpizio, S., Bianchi, C. & Stojanovic, I. Slurs in quarantine. *Mind Lang* **39**, 381–396 (2024).
82. Domaneschi, F., Cepollaro, B. & Stojanovic, I. Literally 'a jerk': an experimental investigation of expressives in predicative position. *Lang Cogn* **17**, e13 (2025).
83. Czopp, A. M., Kay, A. C. & Cheryan, S. Positive Stereotypes Are Pervasive and Powerful. *Perspectives on Psychological Science* **10**, 451–463 (2015).
84. Janschewitz, K. Taboo, emotionally valenced, and emotionally neutral word norms. *Behav Res Methods* **40**, 1065–1074 (2008).
85. Sulpizio, S. *et al.* Taboo language across the globe: A multi-lab study. *Behav Res Methods* **56**, 3794–3813 (2024).
86. Lucisano, P. & Piemontese, M. E. Gulpease: una formula per la predizione della leggibilità di testi in lingua italiana. *SCUOLA E CITTÀ* 110–124 (1998).
87. Honnibal, M., Montani, I., Van Landeghem, S. & Boyd, A. spaCy: industrial-strength natural language processing in Python. Preprint at https://doi.org/https://doi.org/10.5281/zenodo.1212303. (2020).
88. Le Mens, G., Kovács, B., Hannan, M. T. & Pros, G. Uncovering the semantics of concepts using GPT-4. *Proceedings of the National Academy of Sciences* **120**, e2309350120 (2023).
89. Wang, H. *et al.* Prompting Large Language Models for Topic Modeling. in *2023 IEEE International Conference on Big Data (BigData)* 1236–1241 (IEEE, 2023). doi:10.1109/BigData59044.2023.10386113.
90. Gilardi, F., Alizadeh, M. & Kubli, M. ChatGPT outperforms crowd workers for text-annotation tasks. *Proceedings of the National Academy of Sciences* **120**, (2023).
91. Bulatov, V., Alekseev, V. & Vorontsov, K. Determination of the Number of Topics Intrinsically: Is It Possible? https://doi.org/10.1007/978-3-031-67008-4_1 (2024) doi:10.1007/978-3-031-67008-4_1.
92. Jacobs, A. M. The Gutenberg English Poetry Corpus: Exemplary Quantitative Narrative Analyses. *Frontiers in Digital Humanities* **5**, (2018).
93. Xu, X. Interpreting metaphorical statements. *J Pragmat* **42**, 1622–1636 (2010).
94. McGregor, S., Agres, K., Rataj, K., Purver, M. & Wiggins, G. Re-Representing Metaphor: Modeling Metaphor Perception Using Dynamically Contextual Distributional Semantics. *Front Psychol* **10**, 765 (2019).
95. Winter, B. & Strik-Lievers, F. Semantic distance predicts metaphoricity and creativity judgments in synesthetic metaphors. *Metaphor and the Social World* **13**, 59–80 (2023).
96. Stamenković, D., Milenković, K., Ichien, N. & Holyoak, K. J. An Individual-Differences Approach to Poetic Metaphor: Impact of Aptness and Familiarity. *Metaphor Symb* **38**, 149–161 (2023).
97. Garello, S. The visibility of speech. *Pragmatics & Cognition* **30**, 353–376 (2023).
98. Carston, R. Figurative Language, Mental Imagery, and Pragmatics. *Metaphor Symb* **33**, 198–217 (2018).
99. Carston, R. Metaphor: Ad hoc concepts, literal meaning and mental images. *Proceedings of the Aristotelian Society* **110**, 295–321 (2010).

100. Katz, A. N., Paivio, A. & Marschark, M. Poetic comparisons: Psychological dimensions of metaphoric processing. *J Psycholinguist Res* **14**, 365–383 (1985).
101. Pompei, C. *et al.* 'Broccoli is candy': the role of metaphors in children's persuasive communication. *Language Development Research* (2025).
102. Sopory, P. & Dillard, J. P. The Persuasive Effects of Metaphor: A Meta-Analysis. *Hum Commun Res* **28**, 382–419 (2002).
103. Mitić, A. J., Mitić, I. & Stamenković, D. Can reading habits affect metaphor evaluation? Exploring key relations. *Open Linguist* **11**, (2025).
104. Schäffner, C. & Shuttleworth, M. Metaphor in translation. *Target. International Journal of Translation Studies* **25**, 93–106 (2013).
105. Kövecses, Z. Language, Figurative Thought, and Cross-Cultural Comparison. *Metaphor Symb* **18**, 311–320 (2003).
106. Ifantidou, E. & Hatzidaki, A. Metaphor comprehension in L2: Meaning, images and emotions. *J Pragmat* **149**, 78–90 (2019).
107. Kövecses, Z. Metaphor, language, and culture. *DELTA: Documentação de Estudos em Lingüística Teórica e Aplicada* **26**, 739–757 (2010).
108. R Core Team. R: A language and environment for statistical computing. Preprint at (2021).
109. Chang, W. *et al.* shiny: Web Application Framework for R. Preprint at https://CRAN.R-project.org/package=shiny (2022).
110. Chang, W. & Borges Ribeiro, B. shinydashboard: Create Dashboards with 'Shiny'. Preprint at https://CRAN.R-project.org/package=shinydashboard (2021).